\documentclass{article} 
\usepackage[preprint]{colm2026_conference}

\usepackage{microtype}
\usepackage{url}

\usepackage{inconsolata}

\usepackage{graphicx}
\usepackage{mathtools} 
\usepackage{booktabs} 
\usepackage{tabularx} 
\usepackage{subcaption} 
\usepackage{comment}
\usepackage{graphicx}
\usepackage[table]{xcolor} 
\usepackage{array}    
\usepackage{float}
\usepackage{amsmath}
\usepackage{amssymb}
\usepackage{amsfonts}
\usepackage{textcomp}
\usepackage{tikz}
\usetikzlibrary{positioning}
\usepackage{xcolor}
\usepackage{booktabs}
\usepackage{longtable}
\usepackage{makecell}
\usepackage[table]{xcolor}
\usepackage{colortbl}

\usepackage{wrapfig}

\usepackage{lineno}

\definecolor{darkblue}{rgb}{0, 0, 0.5}

\title{Algorithmic Blindness in Large Language Models: A Calibration Study of Performance Prediction}

\author{Sohan Venkatesh, Ashish Mahendran Kurapath \& Tejas Melkote \\
Manipal Institute of Technology Bengaluru\\
\texttt{\{sohan1, ashish, tejas2\}.mitblr2022@learner.manipal.edu} 
}

\begin{document}

\ifcolmsubmission
\linenumbers
\fi

\maketitle

\begin{abstract}
Large language models (LLMs) demonstrate remarkable breadth of knowledge, yet their ability to reason about computational processes remains poorly understood. Closing this gap matters for practitioners who rely on LLMs to guide algorithm selection and deployment. We address this limitation using causal discovery as a testbed and evaluate eight frontier LLMs against ground truth derived from algorithm executions. We find systematic, near-total failure across models. The predicted ranges are far wider than true confidence intervals yet still fail to contain the true algorithmic mean in most cases. Most models perform worse than random guessing. The best model's marginal improvement points to benchmark memorization rather than principled reasoning. We term this failure algorithmic blindness and argue it reflects a fundamental gap between declarative knowledge about algorithms and calibrated procedural prediction.
\end{abstract}

\section{Introduction}\label{sec:intro}
Can large language models predict how well an algorithm will perform on a given problem instance? If so, they could serve as zero-shot algorithm selectors or calibrated uncertainty estimators, reducing the need for costly empirical evaluation. Whether such recommendations carry calibrated quantitative validity or whether LLMs merely pattern match on training text that breaks down under numerical scrutiny, has not been systematically evaluated.

Answering this requires a domain where algorithmic performance is objectively measurable, algorithms are diverse and well documented and benchmark datasets are sufficiently prominent in training corpora to expose whether above-random performance reflects genuine reasoning or memorization. Causal discovery satisfies all of these requirements. It spans multiple algorithmic paradigms with distinct theoretical assumptions, provides standardized metrics with well-defined ground truth and allows controlled synthetic data generation. This combination makes it a principled testbed for probing structure-conditioned generalization rather than surface-level benchmark recall.

We ask whether frontier LLMs can provide calibrated predictions of algorithm performance. We operationalize this via calibrated coverage, defined as the fraction of comparisons where an LLM's predicted range contains the true algorithmic mean from 100 independent runs. We term the failure we uncover \emph{algorithmic blindness}: the inability of LLMs to form calibrated expectations about algorithm performance from problem structure alone. We argue this failure mode is unlikely to be domain specific since causal discovery serves here as a testbed that makes it visible.

This paper makes four contributions. First, we establish causal discovery as a rigorous testbed for evaluating LLM algorithmic reasoning, combining diverse algorithmic families, standardized metrics and established benchmarks with sufficient prior literature to expose memorization. Second, we conduct the first large-scale calibration study of this kind, spanning eight frontier models, thirteen datasets, four algorithms, four metrics and three prompt formulations against ground truth from 5,200 algorithm runs. Third, we demonstrate that frontier LLMs achieve only 15.9\% calibrated coverage, with seven of eight models falling below a random baseline and simple heuristics outperforming most models tested. Fourth, we show through algorithm-specific performance collapse on held-out synthetic datasets that the marginal above-random performance of the best model is more consistent with retrieval of benchmark-associated statistics than with structure-conditioned generalization.

\begin{figure}[t]
\centering
\includegraphics[width=\linewidth, trim=0pt 290pt 220pt 10pt, clip]{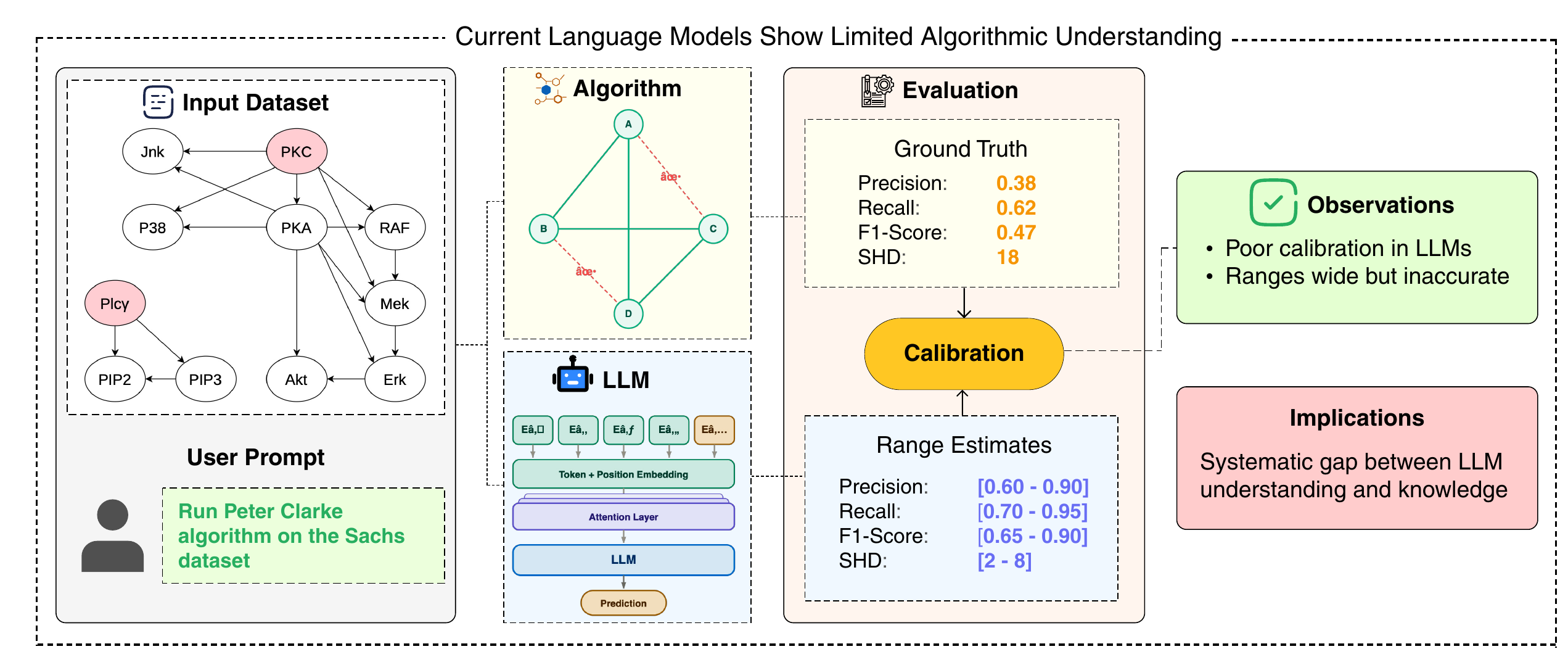}
\caption{Comparison of LLM estimates and algorithmic ground truth revealing algorithmic blindness. The model is prompted to predict performance ranges for the PC algorithm on the Sachs dataset and the predicted ranges are wide yet fail to contain the true values. }
\label{fig:intro}
\end{figure}

\section{Related Work}\label{sec:related-work}

The algorithm selection problem, formalized by \cite{rice1976algorithm}, concerns choosing the best algorithm for a given problem instance based on instance features. Subsequent work developed metalearning approaches that train predictors on algorithm performance histories, enabling informed selection across SAT solvers, combinatorial optimizers and machine learning pipelines. AutoML systems such as Auto-WEKA \citep{thornton2013auto} and Auto-sklearn \citep{feurer2015efficient} extend this paradigm with Bayesian optimization over algorithm configurations.

A growing body of work probes the reasoning capabilities of LLMs. Studies of mathematical and scientific reasoning have found strong surface-level performance that degrades under distribution shift or problem reformulation, suggesting pattern matching over symbolic reasoning \citep{ullman2023large, mirzadeh2024gsm}. Calibration studies show LLMs are systematically overconfident in factual domains \citep{kadavath2022language, xiong2023can}. Our work contributes a specific failure mode: algorithmic performance prediction, where LLM confidence does not translate to predictive validity.

Recent work has explored LLMs as assistants for scientific tasks including experimental design, hypothesis generation and model selection \citep{boiko2023emergent, kambhampati2024position}. LLMs have shown promise in qualitatively ranking algorithmic approaches and suggesting appropriate methods given problem descriptions \citep{tornede2023automl, jiang2024followbench} and \cite{yang2023large} show LLM guided optimization outperforms random baselines in structured search. 

Causal discovery provides structured evaluation criteria (precision, recall, F1 and SHD against a ground-truth DAG), a family of algorithms with well-characterized theoretical properties and a suite of benchmark networks from the bnlearn repository \citep{scutari2010learning} evaluated across hundreds of papers. Prior work has used these benchmarks to compare algorithmic families \cite{heinze2018causal, vowels2022d} and to assess sensitivity of causal methods to assumption violations \citep{kalisch2007estimating}. We are not testing LLMs on causal discovery per se; we are using causal discovery to test whether LLMs can predict algorithmic performance.

Calibration, defined as the alignment between predicted probabilities and empirical frequencies, is central to probabilistic machine learning \citep{guo2017calibration}. We apply this concept to interval prediction: a well-calibrated predictor's stated ranges should contain the true value at the stated rate \citep{gneiting2007strictly}. LLMs have been found to be poorly calibrated in classification settings \citep{kadavath2022language}. Our work simply extends this to interval prediction in a domain requiring deep algorithmic knowledge.

\section{Methodology}\label{sec:methodology}

\subsection{Ground Truth Computation}
For each of 13 datasets combined with 4 algorithms, we run each algorithm 100 times with bootstrap resampling \citep{efron1994introduction}, computing precision, recall, F1 and Structural Hamming Distance (SHD) per run. This yields 5,200 total algorithm executions. From 100 runs per condition, we compute the empirical mean and 95\% confidence interval, which constitute the ground truth for LLM comparison.

\paragraph{Datasets. } We use 9 benchmark datasets from the bnlearn repository\footnote{\url{https://www.bnlearn.com/bnrepository/}} \citep{scutari2010learning}: Alarm \citep{beinlich1989alarm}, Asia \citep{lauritzen1988local}, Cancer \citep{korb2010bayesian}, Child \citep{spiegelhalter1993bayesian}, Earthquake \citep{korb2010bayesian}, Hepar2 \citep{onisko2003probabilistic}, Insurance \citep{binder1997adaptive}, Sachs \citep{sachs2005causal} and Survey \citep{scutari2015bayesian}, ranging from 8 to 70 nodes. These represent standard evaluation benchmarks in the causal discovery literature with high likelihood of presence in LLM training corpora. We additionally construct 4 synthetic datasets with 12, 30, 50 and 60 nodes, generated from random DAGs with controlled Erdős-Rényi edge density \citep{erdHos1960evolution}. Synthetic datasets serve as held-out tests of generalization absent memorizable benchmark statistics.

\paragraph{Algorithms. } We evaluate Peter-Clark (PC) \citep{spirtes2000causation}, Fast Causal Inference (FCI) \citep{richardson2002ancestral}, Linear Non-Gaussian Acyclic Model (LiNGAM) \citep{shimizu2006linear} and Non-combinatorial Optimization via Trace Exponential and Augmented lagRangian for Structure learning (NOTEARS) \citep{zheng2018dags} using standard implementations from the causal-learn library \citep{zheng2024causal}, all with default hyperparameters. Default settings were retained to reflect typical practitioner usage and to avoid conflating performance prediction with dataset-specific hyperparameter tuning.

\paragraph{Metrics. } We evaluate four standard causal discovery metrics \citep{acid2003searching, tsamardinos2006max}. Given a predicted edge set $\hat{E}$ and true edge set $E^*$ over a graph with $d$ nodes, these are defined as:
\begin{equation}
\text{Precision} = \frac{|\hat{E} \cap E^*|}{|\hat{E}|}, \quad \text{Recall} = \frac{|\hat{E} \cap E^*|}{|E^*|}, \quad \text{F1} = \frac{2 \cdot \text{Precision} \cdot \text{Recall}}{\text{Precision} + \text{Recall}}
\end{equation}

Structural Hamming Distance \citep{tsamardinos2006max} counts the minimum number of edge insertions, deletions and direction reversals required to transform the predicted graph into the true DAG:

\begin{equation}
\text{SHD}(\hat{G}, G^*) = |\text{missing edges}| + |\text{extra edges}| + |\text{reversed edges}|
\end{equation}

\subsection{LLM Query Protocol}

We query eight language models: Claude-Opus-4.6 \citep{anthropic2026claude}, GPT-5.2 \citep{openai2026gpt52}, DeepSeek-V3.2-Reasoner \citep{liu2025deepseek}, DeepSeek-R1-0528 \citep{guo2025deepseek}, Qwen3-Next-80B-A3B-Thinking \citep{yang2025qwen3}, Gemini3-Pro-Preview \citep{google2026gemini3}, LLaMA-3.3-70B \citep{meta2026llama33} and Qwen2.5-7B \citep{yang2024qwen2.5}.\footnote{Abbreviated as Claude, GPT-5, DeepSeek-Think, DeepSeek, Qwen-Think, Gemini 3, LLaMA and Qwen respectively for brevity.} Each model is queried with 3 distinct prompt formulations designed to elicit predicted performance ranges across all four metrics. Formulations vary in specificity: direct question with metric names (f1), expanded description with algorithm intuition (f2) and alternative phrasing emphasizing uncertainty (f3). Multiple prompt formulations follow best practices for robust LLM evaluation \citep{mizrahi2024state, sclar2023quantifying}, reducing sensitivity to phrasing choices.

\subsection{Aggregation and Coverage Computation}
To avoid cherry-picking prompt formulations, we average predicted ranges across the three formulations for each model-condition pair, yielding 52 aggregated predictions per model. Calibrated coverage is then computed as the fraction of (model, dataset, algorithm, metric) quadruples where the aggregated predicted range contains the true algorithmic mean. With 8 models × 13 datasets × 4 algorithms × 4 metrics = 1,664 total comparisons, this yields a single primary metric per model and an overall mean across models.

Formally, calibrated coverage is defined as:
\begin{equation}
\text{Coverage}(M, D, A, m) = \mathbf{1}\big[\hat{\mu}_{D,A,m} \in [\hat{l}_{M,D,A,m},\, \hat{u}_{M,D,A,m}]\big]
\end{equation}

where $\hat{\mu}_{D,A,m}$ is the empirical algorithmic mean and $[\hat{l}, \hat{u}]$ is the LLM model $M$'s aggregated predicted range. This metric \citep{gneiting2007strictly} directly answers the operational question of whether an LLM's stated range is informative for a practitioner.

\subsection{Baselines}

We evaluate two baselines on the same 1,664 comparisons. The \emph{random baseline} draws predicted ranges uniformly at random within the valid domain for each metric (e.g., $[0,1]$ for precision, recall and F1; $[0, \text{max\_SHD}]$ for SHD). The \emph{heuristic baseline} constructs ranges using conservative dataset-level statistics from prior literature, with widths scaled to observed algorithm variance. These baselines follow the evaluation practice of establishing uninformed predictors as a reference floor \citep{demvsar2006statistical}.

\subsection{Prompt Robustness and Algorithm-Specific Degradation Analysis}

We compute the coefficient of variation (CV\%) across the three prompt formulations per model-metric-experiment to quantify prompt sensitivity \citep{liang2022holistic}. We additionally analyze algorithm-specific coverage degradation on synthetic versus benchmark datasets by computing, for each algorithm, the mean coverage boost on synthetic data averaged across all 8 models. If degradation is uniform across algorithms, it would indicate a general synthetic-data difficulty effect. Dissociation between heavily benchmarked and less-documented algorithms points to training data coverage rather than genuine algorithmic understanding. Full per-model CV\% data is reported in Appendix~\ref{sec:prompt-form}.

\subsection{Memorization Probes}

We conduct a thorough analysis to probe whether LLM behavior reflects memorization of benchmark statistics. First, we compare predicted range widths on benchmark versus synthetic datasets: under memorization, LLMs should produce tighter ranges for datasets whose statistics they have retrieved and wider ranges for novel synthetic data. Then, we measure cross-model agreement as mean pairwise distance between predicted ranges across models for the same condition. Under memorization, models should converge on benchmark predictions and diverge on synthetic data. Agreement is computed separately for benchmark and synthetic conditions and broken down by metric and network size.

\section{Results}
\subsection{Systematic Coverage Failure}
\label{subsec:primary}

Across all 1,664 comparisons, frontier LLMs achieve a mean calibrated coverage of \textbf{15.9\%}, meaning predicted ranges contain the true algorithmic mean fewer than 1 in 6 times. This represents an 84.1\% failure rate. The 7$\times$ spread between Claude (39.4\%) and Qwen (5.8\%) indicates substantial model-level variance but even the best-performing model achieves coverage far below what would constitute reliable algorithm selection guidance. Table~\ref{tab:model_coverage} reports per-model results.

Failure is not driven by a single algorithm or metric. All four algorithms and all four metrics fall below 21\% coverage (Table~\ref{tab:coverage_marginals}) and the 39\% relative gap between Recall (18.8\%) and Precision (13.5\%) suggests LLMs systematically overestimate true positive rates while underestimating false positives, consistent with retrieving optimistic benchmark summaries rather than reasoning about error structure \citep{xiong2023can}. FCI's 11.3\% coverage, the lowest of any algorithm, reflects a specific difficulty with PAG-structured output and correctness-guarantee reasoning.

\begin{table}[h]
\centering
\footnotesize
\begin{minipage}[c]{0.57\textwidth}
\centering
\setlength{\tabcolsep}{3pt}
\caption{Calibrated Coverage by Model.}
\label{tab:model_coverage}

\begin{tabular}{lccc}
\hline
\rowcolor{gray!20}
\textbf{Model} & \textbf{Cov (\%)} & \textbf{Comparisons} & \textbf{Mean Score} \\
Claude        & 39.4 & 82/208  & 0.442 \\
\rowcolor{gray!10}
GPT-5         & 15.4 & 32/208  & 0.217 \\
DeepSeek-Think& 14.9 & 31/208  & 0.174 \\
DeepSeek      & 14.4 & 30/208  & 0.198 \\
\rowcolor{gray!10}
Qwen-Think    & 13.9 & 29/208  & 0.191 \\
Gemini 3      & 13.0 & 27/208  & 0.182 \\
\rowcolor{gray!10}
LLaMA         & 10.1 & 21/208  & 0.152 \\
Qwen          & 5.8  & 12/208  & 0.068 \\
\hline
\textbf{Mean} & \textbf{15.9} & \textbf{264/1664} & --- \\
\hline
\end{tabular}
\end{minipage}\hfill
\begin{minipage}[c]{0.39\textwidth}
\centering
\scriptsize
\setlength{\tabcolsep}{2pt}
\renewcommand{\arraystretch}{0.95}
\caption{Calibrated Coverage by Algorithm and Metric}
\label{tab:coverage_marginals}

\begin{tabular}{lclc}
\hline
\rowcolor{gray!20}
\textbf{Algorithm} & \textbf{Coverage} & \textbf{Metric} & \textbf{Coverage} \\
NOTEARS & 20.7\% & Recall    & 18.8\% \\
\rowcolor{gray!10}
LiNGAM  & 20.0\% & F1        & 16.3\% \\
PC      & 11.5\% & SHD       & 14.9\% \\
\rowcolor{gray!10}
FCI     & 11.3\% & Precision & 13.5\% \\
\hline
\end{tabular}
\end{minipage}
\end{table}

Claude's above-random performance warrants closer examination. Cross-algorithm breakdown reveals that Claude's synthetic coverage boost is highly algorithm-specific: $+18.8\%$ for FCI, $+24.3\%$ for NOTEARS, $+16.0\%$ for PC but $-16.0\%$ for LiNGAM (Table~\ref{tab:claude_algorithm_breakdown}). This 40.3 percentage point range of variation rules out a general synthetic data difficulty effect. The algorithm-specific nature, with LiNGAM uniquely showing degradation on synthetic data, points to pattern matching against benchmark statistics rather than principled reasoning about algorithm behavior. Claude also shows the strongest range width compression of all models (0.26$\times$ ratio; Table~\ref{tab:range_width}).

\begin{wraptable}{r}{0.52\textwidth}
\centering
\caption{Claude's calibrated coverage by algorithm and dataset type.}
\label{tab:claude_algorithm_breakdown}
\scriptsize
\setlength{\tabcolsep}{2pt}
\begin{tabular}{lccc}
\hline
\rowcolor{gray!20}
\textbf{Algorithm} & \textbf{Benchmark Cov.} & \textbf{Synthetic Cov.} & \textbf{Difference} \\
FCI & 25.0\% & 43.8\% & $+18.8\%$ \\
\rowcolor{gray!10} NOTEARS & 44.4\% & 68.8\% & $+24.3\%$ \\
PC & 27.8\% & 43.8\% & $+16.0\%$ \\
\rowcolor{gray!10} LiNGAM & 47.2\% & 31.2\% & $-16.0\%$ \\
\hline
\end{tabular}
\end{wraptable}
The pattern tracks training data coverage rather than algorithmic properties: LiNGAM has the most extensive benchmark literature and is the only algorithm where Claude's synthetic performance collapses, while NOTEARS, which is newer and less documented, shows the largest synthetic boost. This dissociation, where collapse occurs precisely where benchmark literature is richest and stability where it is sparse, makes explanations based on algorithmic properties unlikely and favors retrieval of memorized statistics as the explanation.

\subsection{Most LLMs Fail to Exceed Random Guessing}

Seven of eight LLMs perform below the random baseline of 36.5\%. The median LLM achieves 13.9\% coverage, less than 40\% of random baseline performance. Claude marginally exceeds random by 2.9 percentage points (39.4\% vs. 36.5\%), a gap not meaningfully distinguishable from chance-level variation and far below practical utility for algorithm selection. The gap between the random baseline and the seven underperforming models is substantial and statistically reliable.\footnote{One-sample binomial test against the null rate of 36.5\%: all seven models below the random baseline yield $p < 0.001$ individually.}

\begin{wraptable}{r}{0.48\textwidth}
\vspace{-6pt}
\centering
\caption{Calibrated Coverage: LLMs versus Baselines}
\label{tab:coverage_baselines}
\scriptsize
\setlength{\tabcolsep}{3pt}
\begin{tabular}{lcc}
\hline
\rowcolor{gray!20}
\textbf{Method} & \textbf{Coverage (\%)} & \textbf{Mean Score} \\
\rowcolor{green!15} Random Baseline & 36.5 & 0.409 \\
Heuristic Baseline & 32.7 & 0.356 \\
\rowcolor{red!15} Claude & 39.4 & 0.442 \\
GPT-5 & 15.4 & 0.217 \\
DeepSeek-Think & 14.9 & 0.174 \\
DeepSeek & 14.4 & 0.198 \\
Qwen-Think & 13.9 & 0.191 \\
Gemini 3 & 13.0 & 0.182 \\
LLaMA & 10.1 & 0.152 \\
Qwen & 5.8 & 0.068 \\
\hline
\end{tabular}
\vspace{-10pt}
\end{wraptable}

This result, shown in Table~\ref{tab:coverage_baselines}, establishes that practitioners would obtain better-calibrated uncertainty estimates from uniformly random interval guessing than from querying 7 of the 8 frontier LLMs tested. The heuristic baseline (32.7\%) similarly outperforms all models except Claude. These findings confirm that LLMs provide no systematic reasoning advantage for algorithm performance prediction \citep{kambhampati2024can}.

LLM predicted ranges are 8 to 27 times wider than true algorithm confidence intervals, yet coverage remains critically low. Wide ranges that still miss the ground truth indicate miscalibrated beliefs \citep{xiong2023can}.

\subsection{Benchmark versus Synthetic Degradation}
\label{sec:benchmarksyn}
\begin{figure}[h]
\centering
\begin{minipage}[t]{0.49\linewidth}
\centering
\includegraphics[width=\linewidth, trim=5 0 5 5, clip]{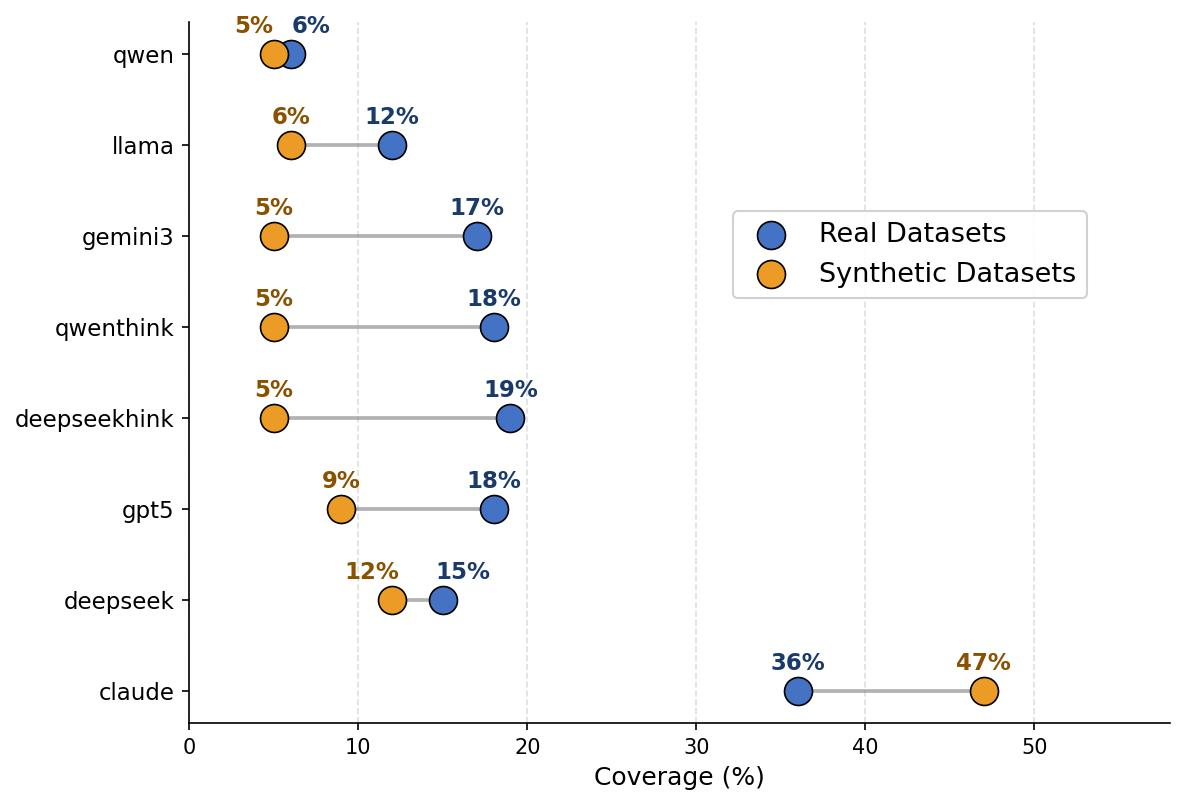}
\caption{Mean calibrated coverage on benchmark datasets versus synthetic datasets across all models.}
\label{fig:real_vs_synthetic}
\end{minipage}\hfill
\begin{minipage}[t]{0.49\linewidth}
\centering
\includegraphics[width=0.80\linewidth, trim=5 5 5 5, clip]{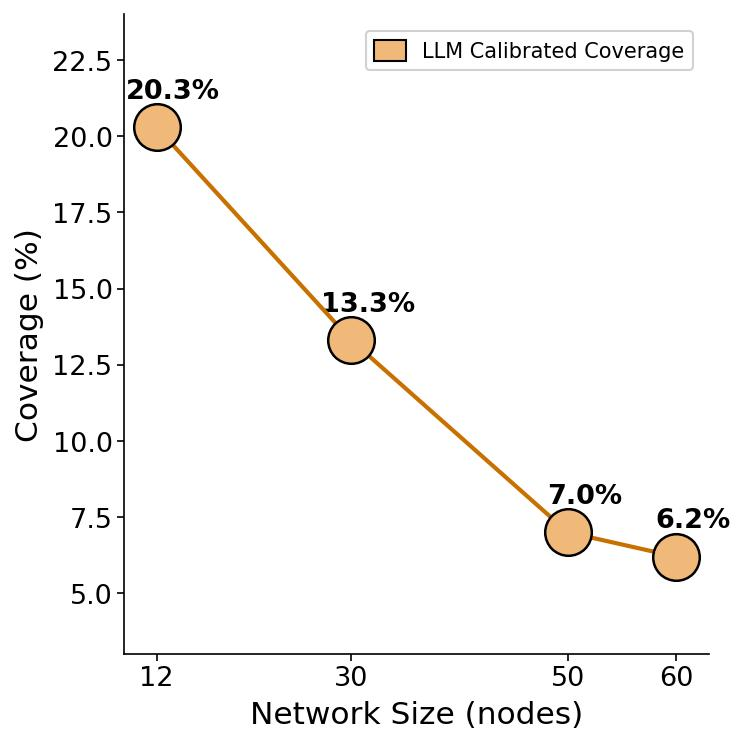}
\caption{Coverage collapse across synthetic networks.}
\label{fig:scalability_collapse}
\end{minipage}
\end{figure}

Coverage on benchmark datasets exceeds synthetic dataset coverage by 51\% (17.7\% vs.\ 11.7\%), as shown in Figure~\ref{fig:real_vs_synthetic}. This aggregate pattern holds for 7 of 8 models; Claude is the exception, showing higher synthetic coverage (47\%) than benchmark coverage (36\%), a reversal addressed in Section~\ref{subsec:primary} as evidence of algorithm-specific pattern matching rather than general synthetic difficulty.

Benchmark datasets are well represented in LLM training corpora through papers, tutorials and documentation \citep{golchin2023time}; synthetic datasets generated for this study are not. If LLMs were performing principled algorithmic reasoning, synthetic performance should be comparable. The observed degradation suggests partial retrieval of benchmark-associated performance statistics.

Coverage on synthetic datasets degrades monotonically with network size: 20.3\% at 12 nodes, 13.3\% at 30 nodes, 7.0\% at 50 nodes and 6.2\% at 60 nodes, showing a 69\% relative decline across the synthetic scale (Figure~\ref{fig:scalability_collapse}). Larger synthetic networks have fewer analogues in training data and this monotonic collapse further supports memorization over generalization \citep{mirzadeh2024gsm}.

\subsection{Memorization Probes}
\label{sec:memorization_probes}

\begin{figure}[h]
\centering
\begin{minipage}[t]{0.52\linewidth}
\vspace{0pt}
\centering
\scriptsize
\setlength{\tabcolsep}{3pt}
\renewcommand{\arraystretch}{0.95}
\captionof{table}{Predicted Range Width by Model}
\label{tab:range_width}
\vspace{2pt}
\begin{tabular}{lccc}
\hline
\rowcolor{gray!20}
\textbf{Model} & \textbf{Bench.} & \textbf{Synth.} & \textbf{Ratio} \\
Claude & 6.12 & 23.77 & 0.26 \\
\rowcolor{gray!10} GPT-5 & 7.65 & 19.20 & 0.40 \\
Qwen & 2.02 & 4.58 & 0.44 \\
\rowcolor{gray!10} DeepSeek & 3.75 & 8.48 & 0.44 \\
DeepSeek-Think & 5.56 & 10.34 & 0.54 \\
\rowcolor{gray!10} LLaMA & 2.85 & 3.78 & 0.75 \\
Qwen-Think & 3.20 & 4.02 & 0.80 \\
\rowcolor{gray!10} Gemini 3 & 3.60 & 4.46 & 0.81 \\
\hline
\end{tabular}
\end{minipage}\hfill
\begin{minipage}[t]{0.44\linewidth}
\vspace{0pt}
\centering
\caption{Predicted range width: benchmark vs.\ synthetic datasets}
\label{fig:variance_benchmark}
\vspace{2pt}
\includegraphics[width=0.75\linewidth, trim=0 40 0 5, clip]{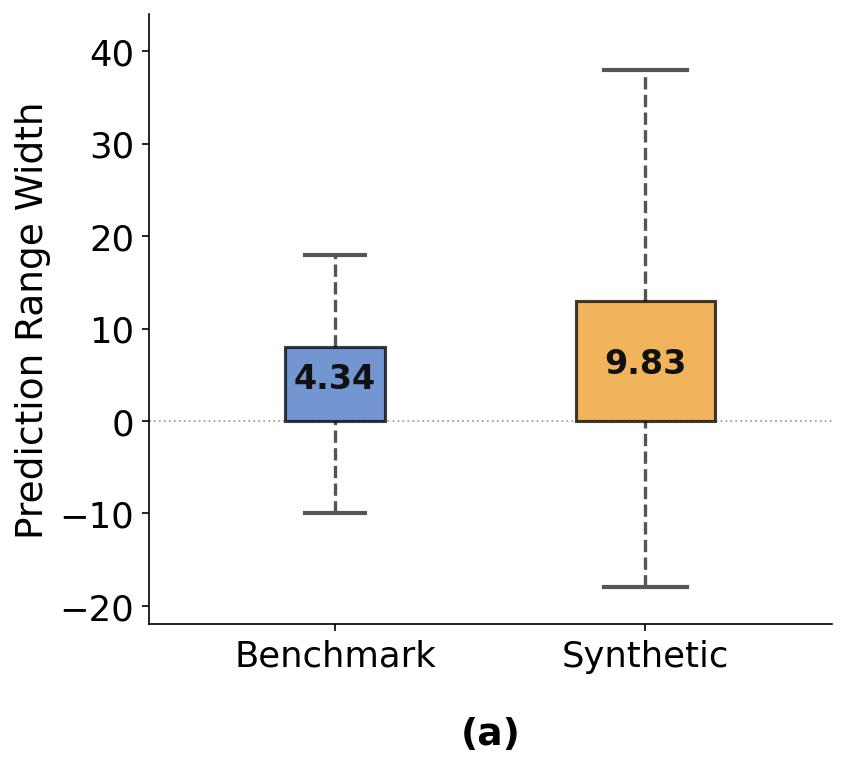}
\end{minipage}
\end{figure}

Under memorization, LLMs should produce tighter ranges for benchmark datasets whose statistics they have retrieved and wider ranges for novel synthetic data. Benchmark datasets yield a mean predicted range width of 4.34, compared to 9.83 for synthetic datasets, giving a 2.26$\times$ compression ratio that holds across all 8 models without exception (Table~\ref{tab:range_width}). The range width also expands monotonically with synthetic network size from 2.55 at 12 nodes to 18.08 at 60 nodes, a 7.1$\times$ increase in a regime where memorization is impossible.

Figure~\ref{fig:variance_benchmark} shows the 2.26$\times$ aggregate compression across benchmark and synthetic datasets. Claude shows the strongest compression (0.26$\times$), which may reflect greater exposure to causal discovery literature during pretraining \citep{kandpal2023large}. Gemini~3 and Qwen-Think show the weakest compression, indicating their benchmark and synthetic ranges are similarly wide. This suggests they treat both dataset types as equally unfamiliar rather than retrieving statistics for known benchmarks.

\begin{wraptable}{r}{0.50\textwidth}
\vspace{-12pt}
\centering
\scriptsize
\setlength{\tabcolsep}{2pt}
\renewcommand{\arraystretch}{0.92}
\caption{Cross-Model Agreement by Dataset (Mean Pairwise Distance and Agreement\%)}
\label{tab:agreement_by_dataset}
\begin{tabular}{lcc>{\centering\arraybackslash}p{1.15cm}}
\hline
\rowcolor{gray!20}
\textbf{Dataset} & \textbf{Type} & \textbf{Mean Dist.} & \textbf{Agree.\%} \\
Asia & Benchmark & 2.20 & 44.2 \\
\rowcolor{gray!10} Cancer & Benchmark & 1.88 & 34.8 \\
Earthquake & Benchmark & 2.01 & 43.5 \\
\rowcolor{gray!10} Survey & Benchmark & 2.13 & 43.3 \\
Sachs & Benchmark & 4.55 & 47.8 \\
\rowcolor{gray!10} Child & Benchmark & 8.62 & 46.2 \\
Alarm & Benchmark & 17.39 & 50.7 \\
\rowcolor{gray!10} Insurance & Benchmark & 17.97 & 44.9 \\
Hepar2 & Benchmark & 59.14 & 42.6 \\
\rowcolor{gray!10} Synthetic-12 & Synthetic & 4.53 & 57.8 \\
Synthetic-30 & Synthetic & 16.77 & 54.0 \\
\rowcolor{gray!10} Synthetic-50 & Synthetic & 42.56 & 52.2 \\
Synthetic-60 & Synthetic & 68.25 & 45.1 \\
\hline
\end{tabular}
\vspace{-10pt}
\end{wraptable}
Cross-model agreement provides the second signal (Table~\ref{tab:agreement_by_dataset}). If models are independently retrieving the same benchmark statistics, they should converge on similar predictions regardless of architecture \citep{carlini2022quantifying}. We find 2.6$\times$ greater pairwise disagreement on synthetic datasets than benchmark datasets (mean pairwise distance 33.03 vs.\ 12.88). Simple well-known benchmarks show the tightest convergence, with Asia and Cancer yielding distances of 2.20 and 1.88, while Synthetic-60 reaches 68.25. The SHD metric is especially revealing: benchmark pairwise distance of 49.99 rises to 130.66 on synthetic data, a sign that models have no principled basis for predicting edge direction accuracy and produce wildly divergent SHD estimates when benchmark figures are unavailable. Agreement collapses 15$\times$ across the synthetic scale, which is inconsistent with structured algorithmic reasoning.

\begin{figure}[h]
\centering
\includegraphics[width=\linewidth]{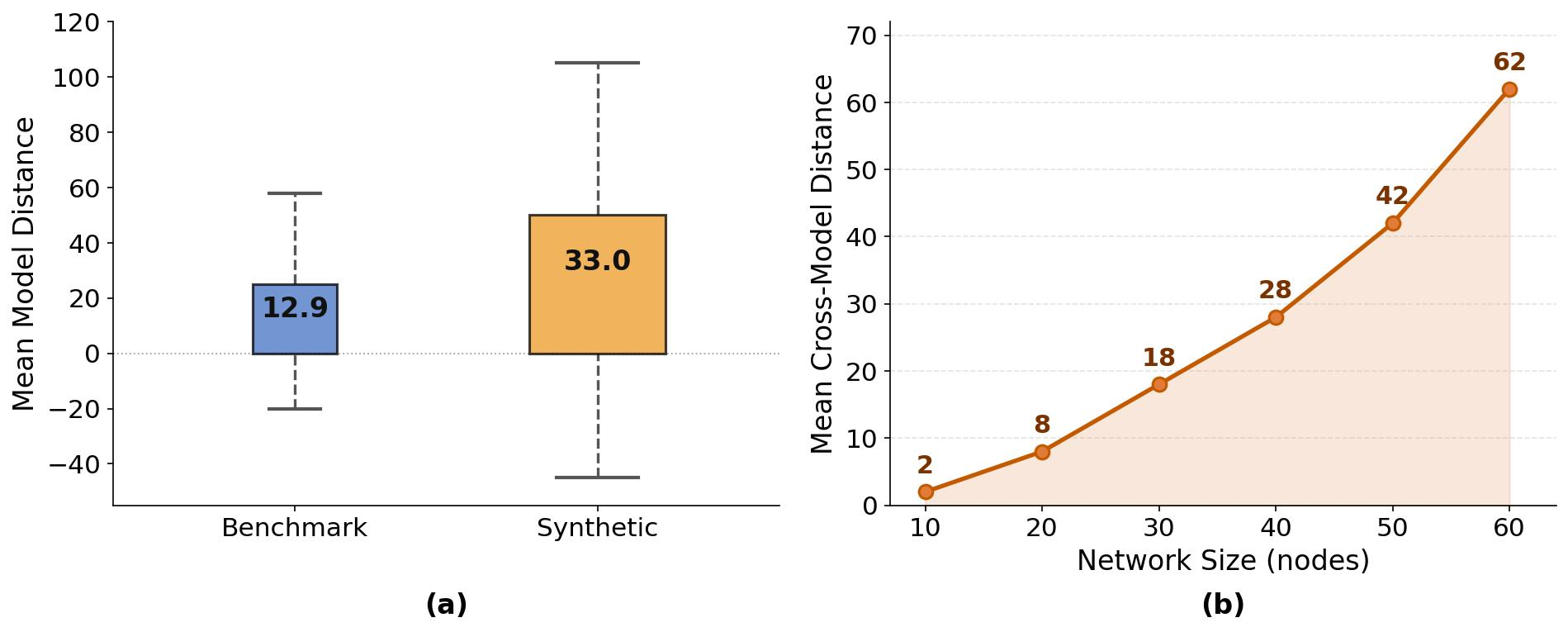}
\caption{(a) Cross-model pairwise distance on benchmark versus synthetic datasets. (b) Mean cross-model pairwise distance as a function of synthetic network size.}
\label{fig:consistency_combined}
\end{figure}

Figure~\ref{fig:consistency_combined}(a) shows the aggregate picture: mean pairwise distance is 12.9 on benchmark datasets and 33.0 on synthetic, a 2.6$\times$ gap that holds across all metric types. Figure~\ref{fig:consistency_combined}(b) resolves the synthetic side by network size: distance rises monotonically from 4.5 at 12 nodes to 68.25 at 60 nodes, a 15$\times$ increase. This monotonic curve is compelling evidence against principled reasoning: if models were reasoning from algorithm and graph properties, larger graphs would not systematically produce more divergent predictions than smaller ones. Instead, the curve reflects models guessing independently as training-data analogues disappear, producing maximal disagreement at the largest network sizes.

\section{Discussion and Limitations}
Taken together, the coverage failure, benchmark-synthetic asymmetry and memorization probes converge on a single picture: LLMs should not be used as zero-shot performance predictors for causal discovery algorithm selection. Predicted ranges are substantially wider than empirical confidence intervals yet still miss the true mean and this degradation scales monotonically with structural complexity. These structured failure modes suggest that the limitation lies not in confidence expression alone but in the absence of a performance model linking algorithmic assumptions to dataset properties. We define \emph{algorithmic blindness} as the failure of a model to translate declarative knowledge of algorithmic assumptions into calibrated expectations about empirical performance from problem structure and algorithmic description alone. This is different from factual ignorance. LLMs may correctly describe algorithmic assumptions yet cannot apply that knowledge to generate accurate performance expectations. This distinction matters for interventions since retrieval-augmented generation or knowledge base integration would address factual gaps but not the absence of a structured performance model identified here.

\paragraph{Limitations. }
Our evaluation covers four algorithms within causal discovery. Algorithmic blindness may manifest differently in domains with different training data coverage or algorithmic diversity. Chain-of-thought, retrieval-augmented generation or specialized system prompts may yield higher coverage. LLM capabilities evolve rapidly and our results characterize frontier models as of the evaluation date, which may not reflect future versions. 

We treat four metrics as independent for coverage computation. In practice, precision, recall, F1 and SHD are correlated, which may affect interpretation of aggregate coverage rates. Ground truth is based on empirical means from 100 bootstrap runs, which may be an unstable target on finite samples for algorithms with high variance. Our memorization inference relies on indirect behavioral signals such as range width compression and cross-model agreement collapse instead of direct training data attribution, so we cannot rule out alternative explanations for these patterns.

\section{Conclusion}

Frontier LLMs exhibit systematic algorithmic blindness when predicting causal discovery algorithm performance. The overwhelming majority of LLM predictions fail to contain the true algorithmic mean and most models perform worse than random guessing. The marginal above-random performance of the sole exception is most consistent with benchmark memorization rather than genuine reasoning. LLM predicted ranges are orders of magnitude wider than true confidence intervals yet remain systematically misaligned with empirical performance. These results establish a clear negative finding for LLM-assisted algorithm selection in causal discovery and motivate careful empirical evaluation before deploying LLMs as performance predictors in any algorithmic domain.

\section{Ethics Statement}
This paper studies the algorithmic understanding of various large language models and highlights the fundamental gap between language models' procedural knowledge and their predictions. We use publicly
available models and APIs through official providers and have adhered to their respective terms of use.

\section{Disclosure of AI usage}
Large language models were used for language editing of the draft and code assistance in this work. The authors are solely responsible for the research idea, experiments design and results discussed in this paper.

\bibliography{colm2026_conference}
\bibliographystyle{colm2026_conference}

\appendix
\section{Causal Discovery Algorithms}\label{sec:causal-dis}
This appendix describes the four algorithms evaluated, their theoretical grounding and the rationale for their inclusion.

\paragraph{Peter-Clark (PC). }

The PC algorithm \citep{spirtes2000causation} is a foundational constraint-based causal discovery method. It operates in two phases: a skeleton phase that removes edges between conditionally independent variables using statistical tests, followed by an orientation phase that applies Meek rules to direct edges. PC assumes causal sufficiency (no hidden confounders), acyclicity and the faithfulness condition. It is included as the canonical representative of constraint-based methods and the most widely benchmarked causal discovery algorithm in the literature.

The skeleton phase removes edge $X_i - X_j$ if there exists a conditioning set $\mathbf{S} \subseteq \mathbf{V} \setminus \{X_i, X_j\}$ such that $X_i \perp\!\!\!\perp X_j \mid \mathbf{S}$. For continuous Gaussian data this is tested via partial correlation, with significance assessed using Fisher's $z$-transform:

\begin{equation}
z_{ij \mid \mathbf{S}} =
\frac{1}{2}
\ln
\frac{1 + \hat{\rho}_{ij \mid \mathbf{S}}}
     {1 - \hat{\rho}_{ij \mid \mathbf{S}}}
\cdot
\sqrt{n - |\mathbf{S}| - 3}
\end{equation}

where $\hat{\rho}_{ij \mid \mathbf{S}}$ is the sample partial correlation and $n$ is the sample size. The edge is removed when $|z_{ij \mid \mathbf{S}}|$ falls below the threshold for the chosen significance level. The orientation phase applies Meek's four deterministic rules to direct as many skeleton edges as possible without introducing new v-structures or cycles.

\paragraph{Fast Causal Inference (FCI). }

The FCI algorithm \citep{richardson2002ancestral} extends PC to handle latent confounders and selection bias. Rather than outputting a DAG, FCI produces a Partial Ancestral Graph (PAG) that encodes uncertainty over causal structure in the presence of hidden variables. FCI is included as the constraint-based method that relaxes PC's causal sufficiency assumption, making it more applicable to real-world settings where unmeasured confounders are plausible. Its output is more conservative (less edge-committing) than PC by design.

FCI uses the same conditional independence test as PC for skeleton construction. The key distinction is in the orientation rules: FCI uses a superset of PC's orientation rules, adding rules that propagate edge marks (tail $-$, arrowhead $>$ or circle $\circ$) through the PAG. An edge $X_i \circ\!\!-\!\!\circ X_j$ in the initial skeleton is oriented to $X_i \circ\!\!\rightarrow X_j$ when $X_i$ is found to be in the Markov boundary of $X_j$ but not vice versa under the ancestral graph constraints. The resulting PAG represents an equivalence class of MAGs (Maximal Ancestral Graphs), where each edge mark encodes what is invariant across all causal structures consistent with the conditional independence constraints.

\paragraph{Linear Non-Gaussian Acyclic Model (LiNGAM).}

The LiNGAM algorithm \citep{shimizu2006linear} exploits non-Gaussianity of error terms to achieve full identifiability of the causal DAG from observational data alone, a result impossible under the Gaussian assumption. LiNGAM uses Independent Component Analysis to recover the causal ordering and edge weights. It is included as the representative functional causal model and as a contrast to constraint-based methods: its identifiability guarantee and ICA-based mechanism produce qualitatively different performance characteristics across datasets. LiNGAM performs strongly on datasets with genuinely non-Gaussian noise (e.g., Survey) but degrades on datasets that violate its linear assumptions.

LiNGAM assumes the data-generating process follows the structural equation model:

\begin{equation}
\mathbf{x} = \mathbf{B}\mathbf{x} + \mathbf{e}
\end{equation}

where $\mathbf{x} \in \mathbb{R}^d$ is the observed variable vector, $\mathbf{B}$ is a strictly lower-triangular weighted adjacency matrix (encoding the DAG) and $\mathbf{e}$ is a vector of mutually independent non-Gaussian noise terms. Rearranging gives

\begin{equation}
\mathbf{x} = (\mathbf{I} - \mathbf{B})^{-1}\mathbf{e} = \mathbf{A}\mathbf{e},
\end{equation}

which is an ICA model. The algorithm recovers $\mathbf{A}$ via ICA, then finds the permutation matrix $\mathbf{P}$ such that $\mathbf{P}\mathbf{A}^{-1}$ is strictly lower-triangular, yielding the causal ordering and edge weights.

\paragraph{Non-combinatorial Optimization via Trace Exponential and Augmented lagRangian for Structure learning (NOTEARS). }

NOTEARS \citep{zheng2018dags} reformulates the combinatorial problem of DAG structure learning as a continuous optimization problem using a smooth acyclicity constraint.

\begin{equation}
h(\mathbf{W}) =
\operatorname{tr}\!\left(
\exp\{\mathbf{W} \circ \mathbf{W}\}
\right)
- d = 0
\end{equation}

This allows gradient-based optimization over the space of weighted adjacency matrices. NOTEARS is included as the representative continuous optimization approach, which differs fundamentally in algorithmic mechanism from both constraint-based and functional methods. Its performance is strongest on synthetic data generated from linear Gaussian models (which match its optimization objective) and weaker on benchmark networks with complex nonlinear structure.

The full optimization problem is:
\begin{equation}
\min_{\mathbf{W} \in \mathbb{R}^{d \times d}} \;
\frac{1}{2n}
\left\|
\mathbf{X} - \mathbf{X}\mathbf{W}^T
\right\|_F^2 \\
\text{subject to} \\
h(\mathbf{W}) =
\operatorname{tr}\!\left(
\exp\{\mathbf{W} \circ \mathbf{W}\}
\right)
- d = 0
\end{equation}

where $\mathbf{X} \in \mathbb{R}^{n \times d}$ is the data matrix, $\mathbf{W}$ is the weighted adjacency matrix of the learned DAG, $\circ$ denotes the elementwise product and $h(\mathbf{W}) = 0$ is satisfied if and only if $\mathbf{W}$ is acyclic. The constraint is solved via an augmented Lagrangian method, converting the constrained problem into a sequence of unconstrained subproblems amenable to standard gradient-based optimizers.

\paragraph{Rationale for Algorithm Selection. }

The four algorithms span the three major algorithmic families in causal discovery: constraint-based methods (PC, FCI), functional causal models (LiNGAM) and continuous optimization (NOTEARS). This coverage ensures that results are not specific to one algorithmic paradigm. The algorithms also differ in their theoretical assumptions (causal sufficiency, linearity, Gaussianity), output type (DAG vs.\ PAG vs.\ weighted matrix) and sensitivity to dataset characteristics, making algorithm performance prediction a genuinely nontrivial task that requires understanding the interaction between algorithmic properties and dataset structure.

\section{Prompt Formulations and CV\% analysis}
\label{sec:prompt-form}

\begin{figure}[h]
\centering
\resizebox{\textwidth}{!}{%
\begin{tikzpicture}

\fill[blue!5] (0.2,0.2) rectangle (5.9,9.8);
\draw[blue!60!black, line width=0.8pt, rounded corners=3pt] (0.2,0.2) rectangle (5.9,9.8);

\node[font=\bfseries\normalsize, text=blue!70!black, anchor=north] 
at (3.05,9.6) {Formulation 1: Direct};

\node[
    font=\sffamily\footnotesize,
    text width=5.3cm,
    anchor=north west,
    align=left
] at (0.35,8.8) {%
\textit{You are an expert in causal}\\
\textit{discovery algorithms.}\\[3pt]
Dataset: \texttt{<dataset\_name>}\\
Variables: \texttt{<n\_nodes>}\\
Samples: \texttt{<n\_samples>}\\
Data type: \texttt{<data\_type>}\\[3pt]
Algorithm: \texttt{<algorithm\_name>}\\[3pt]
Estimate performance ranges\\
for all four metrics:\\[3pt]
Precision: [X.XX, X.XX]\\
Recall: [X.XX, X.XX]\\
F1: [X.XX, X.XX]\\
SHD: [X, X]\\[5pt]
\textbf{CRITICAL:} Output ONLY\\
these four lines. No reasoning,\\
no explanations, no preamble.
};

\fill[green!5] (6.1,0.2) rectangle (11.8,9.8);
\draw[green!60!black, line width=0.8pt, rounded corners=3pt] (6.1,0.2) rectangle (11.8,9.8);

\node[font=\bfseries\normalsize, text=green!60!black, anchor=north] 
at (8.95,9.6) {Formulation 2: Step-by-Step};

\node[
    font=\sffamily\footnotesize,
    text width=5.3cm,
    anchor=north west,
    align=left
] at (6.25,8.8) {%
\textit{You are an expert in causal}\\
\textit{discovery algorithms.}\\[3pt]
Dataset: \texttt{<dataset\_name>}\\
Variables: \texttt{<n\_nodes>}\\
Samples: \texttt{<n\_samples>}\\
Complexity: \texttt{<complexity>}\\[3pt]
Algorithm: \texttt{<algorithm\_name>}\\[3pt]
Before predicting, reason through:\\[3pt]
Core assumptions of the algorithm?\\
Does the dataset satisfy these assumptions?\\
How does complexity affect reliability?\\
What range is realistic?\\[5pt]
\textbf{CRITICAL:} After reasoning, output ONLY:\\[3pt]
Precision: [X.XX, X.XX]\\
Recall: [X.XX, X.XX]\\
F1: [X.XX, X.XX]\\
SHD: [X, X]
};

\fill[orange!5] (12.1,0.2) rectangle (17.8,9.8);
\draw[orange!80!black, line width=0.8pt, rounded corners=3pt] (12.1,0.2) rectangle (17.8,9.8);

\node[font=\bfseries\normalsize, text=orange!80!black, anchor=north] 
at (14.95,9.6) {Formulation 3: Meta-Knowledge};

\node[
    font=\sffamily\footnotesize,
    text width=5.3cm,
    anchor=north west,
    align=left
] at (12.25,8.8) {%
\textit{You are a statistician evaluating}\\
\textit{causal discovery algorithms.}\\[3pt]
A researcher repeatedly runs\\
\texttt{<algorithm\_name>} on\\
\texttt{<dataset\_name>} with different\\
random seeds.\\[3pt]
Dataset characteristics:\\
Variables: \texttt{<n\_nodes>}\\
Samples: \texttt{<n\_samples>}\\
Data type: \texttt{<data\_type>}\\[3pt]
What ranges capture 95\% of\\
typical outcomes?\\[5pt]
\textbf{CRITICAL:} Output ONLY:\\[3pt]
Precision: [X.XX, X.XX]\\
Recall: [X.XX, X.XX]\\
F1: [X.XX, X.XX]\\
SHD: [X, X]\\[3pt]
};

\end{tikzpicture}
}
\caption{Three prompt formulations used across all experimental conditions. Formulation 1 elicits direct numerical estimates; Formulation 2 guides explicit step-by-step reasoning about algorithm assumptions; Formulation 3 frames the task as confidence interval estimation over repeated runs. All three require identical output format to enable CV\% comparison.}
\label{fig:prompt_formulations}
\end{figure}
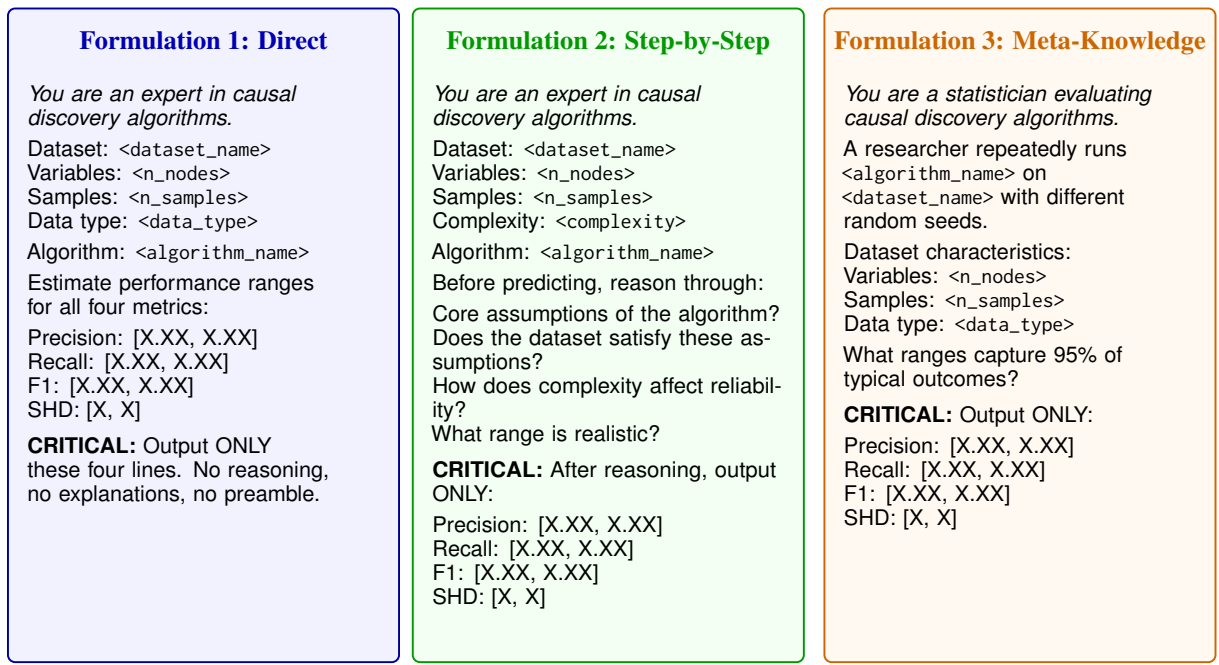

We evaluate prompt robustness by querying each model with three distinct formulations per experimental condition and computing the coefficient of variation (CV\%) of predicted range midpoints and widths across formulations. The three formulations vary along two dimensions: (1) specificity of metric naming, where f1 uses direct metric names, f2 adds algorithm intuition and context and f3 uses alternative uncertainty-focused phrasing; and (2) framing of the prediction task, ranging from direct numerical elicitation to open-ended range description. Figure~\ref{fig:prompt_formulations} shows the three prompt formulations used across all experimental conditions.

\paragraph{CV\% Formula. }

For a given model-metric-experiment triple, let $x_1, x_2, x_3$ denote the predicted midpoints (or widths) across the three formulations. Then:
\begin{equation}
\begin{aligned}
\text{CV\%} &= \frac{\sigma}{\bar{x}} \times 100, \quad
\bar{x} = \frac{1}{3}\sum_{i=1}^{3} x_i, \\
\sigma &=
\sqrt{
\frac{1}{3}
\sum_{i=1}^{3}
(x_i - \bar{x})^2
}
\end{aligned}
\end{equation}

CV\% is computed separately for range midpoints and range widths. High CV\% indicates predictions are sensitive to prompt phrasing; CV\% = 0 indicates identical predictions across all three formulations.

\paragraph{Per-Model CV\% by Metric. }

Average CV\% (midpoint) across all dataset-algorithm combinations within each metric. SHD is an integer-valued graph edit distance; F1, Precision and Recall are bounded continuous scores. Detailed results are shown in table~\ref{tab:prompt_cv}

\begin{table*}[t]
\centering
\caption{Prompt Robustness (CV\%) by Metric and Model}
\label{tab:prompt_cv}
\small
\setlength{\tabcolsep}{4pt}
\begin{tabular}{lcccccccc}
\hline
\rowcolor{gray!20}
\textbf{Metric} & \textbf{Claude} & \textbf{DeepSeek} & \textbf{DeepSeek-Think} & \textbf{Gemini 3} & \textbf{GPT-5} & \textbf{LLaMA} & \textbf{Qwen} & \textbf{Qwen-Think} \\
F1 & 17.1\% & 18.3\% & 14.0\% & 13.9\% & 13.9\% & 7.5\% & 5.3\% & 23.0\% \\
\rowcolor{gray!10}
Precision & 15.4\% & 19.6\% & 15.3\% & 13.9\% & 11.9\% & 6.4\% & 4.1\% & 22.7\% \\
Recall & 18.5\% & 18.8\% & 14.0\% & 15.4\% & 15.8\% & 8.3\% & 6.2\% & 23.2\% \\
\rowcolor{gray!10}
SHD & 19.4\% & 41.5\% & 36.4\% & 23.0\% & 26.5\% & 37.1\% & 30.4\% & 35.1\% \\
\textbf{Overall} & \textbf{17.6\%} & \textbf{24.6\%} & \textbf{19.9\%} & \textbf{16.5\%} & \textbf{17.0\%} & \textbf{14.8\%} & \textbf{11.5\%} & \textbf{26.0\%} \\
\hline
\end{tabular}
\end{table*}

\paragraph{Interpretation of CV\% Patterns. }

Midpoint CV\% ranges from 2.7\% to 151.1\% and width CV\% from 0.0\% to 50.8\% across all model-condition pairs. Maxima exceeding 100\% occur primarily in conditions where predicted midpoints are near zero (e.g., precision or recall on large synthetic graphs), where small absolute differences produce large relative variation; these cases should be interpreted with this scale-dependence in mind. Notably, the models with the highest average CV\%, DeepSeek variants and Qwen-Think, are those explicitly designed for reasoning or multi-step thinking. This pattern suggests that reasoning-focused architectures may be more susceptible to prompt sensitivity, perhaps because their extended inference pathways amplify minor differences in initial phrasing. Conversely, the smallest model (Qwen) shows the lowest CV\%, indicating more consistent (though not more accurate) predictions.

\paragraph{Implications for Main Results. }

Per-model averages of 11 to 26\% reflect genuine phrasing sensitivity across the full range of conditions. However, high sensitivity does not imply that a different prompt set would yield systematically higher coverage. Examination of individual formulation performance reveals that no single formulation consistently outperforms others across models or conditions; sensitivity manifests as instability rather than latent capability being unlocked by specific phrasing \citep{mizrahi2024state, sclar2023quantifying}. Aggregating across all three formulations before computing calibrated coverage, using the mean of f1, f2, f3 lower and upper bounds respectively, ensures the primary findings reflect consistent behavior across varied elicitation strategies rather than any single prompt's characteristics. The 15.9\% coverage result is therefore robust to prompt variation and the observed sensitivity itself reinforces the conclusion that LLMs lack stable, calibrated beliefs about algorithm performance.

\section{Algorithm versus LLM comparison}
This appendix illustrates the comparison structure with two representative datasets: Asia (benchmark, 8 nodes, highest coverage at 23.4\%) and Synthetic-12 (synthetic, 12 nodes, highest synthetic coverage at 20.3\%). Each table shows the algorithmic ground truth mean, the mean LLM predicted range averaged across all 8 models and the percentage of models whose range contained the true mean. Algo Mean is the empirical mean over 100 algorithm runs. LLM Range is the mean predicted interval averaged across all 8 models' aggregated predictions. Coverage is the percentage of the 8 models whose predicted range contained the true algorithmic mean (multiples of 12.5\%).

\begin{table}[h]
\centering
\caption{Calibrated Coverage by Algorithm and Metric for Asia and Synthetic-12 datasets}
\small
\setlength{\tabcolsep}{3pt}

\rowcolors{2}{gray!10}{white}
\begin{tabular}{llccc|llccc}
\toprule
\rowcolor{gray!20}
\multicolumn{5}{c|}{\textbf{Asia (Benchmark, 8 nodes)}} &
\multicolumn{5}{c}{\textbf{Synthetic-12 (Synthetic, 12 nodes)}} \\

\rowcolor{gray!20}
\textbf{Algorithm} & \textbf{Metric} & \textbf{Algo Mean} & \textbf{LLM Range} & \textbf{Coverage} &
\textbf{Algorithm} & \textbf{Metric} & \textbf{Algo Mean} & \textbf{LLM Range} & \textbf{Coverage} \\
\midrule

PC & Precision & 0.474 & [0.703, 0.874] & 12.5\% &
PC & Precision & 0.470 & [0.688, 0.860] & 12.5\% \\

PC & Recall & 0.777 & [0.626, 0.811] & 87.5\% &
PC & Recall & 0.723 & [0.613, 0.797] & 75.0\% \\

PC & F1 & 0.588 & [0.664, 0.831] & 0.0\% &
PC & F1 & 0.569 & [0.650, 0.815] & 0.0\% \\

PC & SHD & 15.0 & [2.1, 6.2] & 0.0\% &
PC & SHD & 32.8 & [4.9, 13.6] & 0.0\% \\

\midrule

FCI & Precision & 0.474 & [0.666, 0.853] & 12.5\% &
FCI & Precision & 0.472 & [0.616, 0.805] & 12.5\% \\

FCI & Recall & 0.777 & [0.588, 0.788] & 62.5\% &
FCI & Recall & 0.722 & [0.549, 0.753] & 75.0\% \\

FCI & F1 & 0.588 & [0.625, 0.807] & 12.5\% &
FCI & F1 & 0.571 & [0.577, 0.765] & 12.5\% \\

FCI & SHD & 15.0 & [2.9, 7.8] & 0.0\% &
FCI & SHD & 32.6 & [8.5, 20.4] & 0.0\% \\

\midrule

LiNGAM & Precision & 0.264 & [0.299, 0.488] & 25.0\% &
LiNGAM & Precision & 0.256 & [0.750, 0.905] & 0.0\% \\

LiNGAM & Recall & 0.362 & [0.265, 0.461] & 37.5\% &
LiNGAM & Recall & 0.455 & [0.702, 0.878] & 12.5\% \\

LiNGAM & F1 & 0.305 & [0.281, 0.459] & 37.5\% &
LiNGAM & F1 & 0.326 & [0.726, 0.883] & 0.0\% \\

LiNGAM & SHD & 13.3 & [7.2, 13.4] & 50.0\% &
LiNGAM & SHD & 33.2 & [4.0, 12.9] & 0.0\% \\

\midrule

NOTEARS & Precision & 0.246 & [0.567, 0.773] & 0.0\% &
NOTEARS & Precision & 0.953 & [0.746, 0.908] & 25.0\% \\

NOTEARS & Recall & 0.250 & [0.510, 0.729] & 25.0\% &
NOTEARS & Recall & 0.456 & [0.696, 0.868] & 12.5\% \\

NOTEARS & F1 & 0.248 & [0.540, 0.732] & 0.0\% &
NOTEARS & F1 & 0.615 & [0.724, 0.880] & 12.5\% \\

NOTEARS & SHD & 12.1 & [3.9, 9.5] & 12.5\% &
NOTEARS & SHD & 10.2 & [4.5, 13.6] & 75.0\% \\

\bottomrule
\end{tabular}
\end{table}

Comparing the two tables illustrates key patterns from the main results. On Asia, PC Recall achieves 87.5\% coverage (the highest single combination in the study) while SHD coverage is 0\% for PC and FCI. On Synthetic-12, LiNGAM collapses to 0\% across precision, F1 and SHD, while NOTEARS SHD reaches 75\%, directly reflecting the algorithm$\times$metric dissociation discussed more in Appendix~\ref{app:algo}.

Two patterns are visible in Table~\ref{tab:calibrated_dataset}. First, three of the four synthetic datasets rank among the four lowest positions, aligning with the benchmark/synthetic degradation reported in Section~\ref{sec:benchmarksyn}. Second, within benchmark datasets, coverage correlates inversely with network size: Asia (8 nodes, 23.4\%) and Cancer (5 nodes, 21.9\%) are the highest-performing benchmarks, while Hepar2 (70 nodes, 10.9\%) is the lowest. Synthetic-12 is the sole exception, achieving 20.3\% coverage comparable to small benchmarks, consistent with its network size falling within the range of well-documented benchmark graphs. The monotonic synthetic collapse from 20.3\% to 6.2\% across 12 to 60 nodes is reported separately in Section~\ref{sec:benchmarksyn}.

\begin{table}[h]
\centering
\small
\captionsetup{skip=2pt}
\caption{Calibrated Coverage by Dataset (Aggregated Across All Algorithms and Metrics)}
\label{tab:calibrated_dataset}
\vspace{4pt}

\begin{tabular}{lccc}
\hline
\rowcolor{gray!20}
\textbf{Dataset} & \textbf{Type} & \textbf{Nodes} & \textbf{Coverage (\%)} \\
\rowcolor{gray!10} Asia & Benchmark & 8 & 23.4 \\
Cancer & Benchmark & 5 & 21.9 \\
\rowcolor{gray!10} Synthetic-12 & Synthetic & 12 & 20.3 \\
Alarm & Benchmark & 37 & 18.8 \\
\rowcolor{gray!10} Insurance & Benchmark & 27 & 18.8 \\
Survey & Benchmark & 6 & 18.0 \\
\rowcolor{gray!10} Child & Benchmark & 20 & 17.2 \\
Sachs & Benchmark & 11 & 16.4 \\
\rowcolor{gray!10} Earthquake & Benchmark & 5 & 14.1 \\
Synthetic-30 & Synthetic & 30 & 13.3 \\
\rowcolor{gray!10} Hepar2 & Benchmark & 70 & 10.9 \\
Synthetic-50 & Synthetic & 50 & 7.0 \\
\rowcolor{gray!10} Synthetic-60 & Synthetic & 60 & 6.2 \\
\hline
\end{tabular}
\end{table}

\section{Algorithm-specific degradation}
\label{app:algo}

The synthetic coverage drop is not uniform across algorithms, which rules out a general synthetic-data difficulty effect. Table~\ref{tab:synthetic_boost_algorithm} reports the average synthetic coverage boost per algorithm across all 8 models.

\begin{wraptable}{r}{0.45\textwidth}
\vspace{-12pt}
\centering
\scriptsize
\setlength{\tabcolsep}{3pt}
\renewcommand{\arraystretch}{0.95}
\caption{Average Synthetic Coverage Boost by Algorithm for all models.}
\label{tab:synthetic_boost_algorithm}

\begin{tabular}{lcc}
\hline
\rowcolor{gray!20}
\textbf{Algorithm} & \textbf{Avg Synthetic Boost} & \textbf{Range Variation} \\
\rowcolor{green!15} NOTEARS & $+1.7\%$ & 56.9\% \\
FCI & $-0.5\%$ & 34.7\% \\
PC & $-2.0\%$ & 24.3\% \\
\rowcolor{red!15} LiNGAM & $-23.2\%$ & 44.4\% \\
\hline
\end{tabular}
\end{wraptable}

LiNGAM collapses by 23.2\% on synthetic data averaged across all models, while NOTEARS shows no degradation ($+1.7\%$). LiNGAM has the most extensive benchmark literature of the four algorithms; NOTEARS is newer with less documented benchmark performance. This dissociation points more toward differential training data exposure than toward principled reasoning. PC and FCI also show modest declines, consistent with partial benchmark memorization.

Crucially, if LLMs had internalized a principled model of algorithmic behavior, degradation on synthetic data should reflect genuine algorithmic properties, since LiNGAM's linear non-Gaussianity assumption is no harder to reason about on synthetic graphs than on benchmarks. Instead, the degradation tracks training data coverage: models perform worse precisely where benchmark statistics are absent, not where the algorithm is intrinsically harder. This suggests that LLM predictions owe more to retrieval of memorized performance figures than to any underlying understanding of how algorithms behave on new data distributions.

\section{Algorithmic Ground Truth and Statistical Analysis}
\paragraph{Algorithm Performance Summary. }
Table~\ref{tab:algo_ground_truth_full} reports the mean Precision, Recall, F1, and SHD for each algorithm, averaged across all 13 datasets with 100 bootstrap runs per dataset–algorithm pair. These values constitute the algorithmic ground truth against which all LLM range predictions are evaluated

\begin{table}[h]
\centering
\small
\captionsetup{skip=2pt}
\caption{Algorithmic Ground Truth: Mean Performance Across 13 Datasets}
\label{tab:algo_ground_truth_full}
\vspace{4pt}

\begin{tabular}{lcccc}
\hline
\rowcolor{gray!20}
\textbf{Algorithm} & \textbf{Precision} & \textbf{Recall} & \textbf{F1} & \textbf{SHD} \\
\rowcolor{gray!10}
PC       & 0.335 & \textbf{0.868} & \textbf{0.427} & 130.36 \\
FCI      & \textbf{0.343} & 0.839 & 0.426 & 122.67 \\
\rowcolor{gray!10}
LiNGAM   & 0.304 & 0.405 & 0.301 & 101.00 \\
NOTEARS  & 0.341 & 0.248 & 0.299 & \textbf{36.77} \\
\hline
\end{tabular}
\end{table}

PC and FCI achieve nearly identical mean F1 (0.427 vs 0.426), reflecting their shared constraint-based approach and similar performance profiles across benchmark datasets. LiNGAM and NOTEARS perform substantially lower on average, though LiNGAM achieves the highest single-dataset F1 (Survey, 0.724) and NOTEARS performs best on synthetic data, consistent with its continuous optimization formulation being less dependent on benchmark-specific graph properties.

\paragraph{Pairwise Significance Testing. }

Table~\ref{tab:bonferroni} reports Bonferroni-corrected pairwise significance tests between algorithms on mean F1 score across all datasets.

\begin{table}[h]
\centering
\small
\caption{Pairwise Algorithm Comparisons (Bonferroni-corrected)}
\label{tab:bonferroni}
\vspace{4pt}
\begin{tabular}{lcccc}
\hline
\rowcolor{gray!20}
\textbf{Comparison} & \textbf{Mean Difference} & \textbf{Corrected p-value} & \textbf{Cohen's d} & \textbf{Significant} \\
\rowcolor{gray!10} FCI vs LiNGAM & +0.125 & 0.013 & 0.842 & Yes \\
LiNGAM vs PC & -0.126 & 0.010 & -0.877 & Yes \\
\rowcolor{gray!10} FCI vs NOTEARS & +0.127 & 1.000 & 0.489 & No \\
FCI vs PC & -0.002 & 1.000 & -0.132 & No \\
\rowcolor{gray!10} LiNGAM vs NOTEARS & +0.003 & 1.000 & 0.009 & No \\
NOTEARS vs PC & -0.128 & 1.000 & -0.495 & No \\
\hline
\end{tabular}
\end{table}

Two of six pairwise comparisons reach significance after correction: FCI vs LiNGAM and LiNGAM vs PC, both with large effect sizes (Cohen's d $>$ 0.84). The non-significance of FCI vs PC confirms that the two constraint-based methods are statistically indistinguishable in ground truth performance, making their divergent LLM coverage (11.3\% vs 11.5\%) unsurprising. The non-significance of NOTEARS vs PC despite a 9.2 percentage point coverage gap (20.7\% vs 11.5\%) further supports the view that LLM coverage differences across algorithms owe more to training data exposure than to true performance differences.

\end{document}